\documentclass[12pt,a4paper]{article}               
\usepackage{listings}
\usepackage{techrep_icg}
\usepackage{amsmath}
\usepackage{caption}
\usepackage{subcaption}
\begin{document}
\reportnr{ICG–CVARLab-TR–003}               
\title{Monocular LSD-SLAM integration within AR System} 
\subtitle{Building a system similar to the HoloLens by integrating LSD-SLAM into our AR Oculus Rift stereo engine for realworld camera tracking} 
\repcity{Graz}            
\repdate{\today}          
\author{Markus H\"oll}
\author[ICG]{Vincent Lepetit}
\newcommand{\TUGn}{Graz University of Technology}
\address[ICG]{Inst. for Computer Graphics and Vision \\ \TUGn, Austria}
\contact{Markus H\"oll}
\contactemail{mhoell@student.tugraz.at}

\begin{abstract}
In this paper, we cover the process of integrating Large-Scale Direct Simultaneous Localization and Mapping (LSD-SLAM) \cite{engel2014lsd} algorithm into our existing AR stereo engine, developed for our modified ''Augmented Reality Oculus Rift''. With that, we are able to track one of our realworld cameras which are mounted on the rift, within a complete unknown environment. This makes it possible to achieve a constant and full augmentation, synchronizing our 3D movement (x, y, z) in both worlds, the real world and the virtual world. The development for the basic AR setup using the Oculus Rift DK1 and two fisheye cameras is fully documented in our previous paper \cite{holl2016augmented}. After an introduction to image-based registration, we detail the LSD-SLAM algorithm and document our code implementing our integration. The AR stereo engine with Oculus Rift support can be accessed via the GIT repository \url{https://github.com/MaXvanHeLL/ARift.git} and the modified LSD-SLAM project used for the integration is available here \url{https://github.com/MaXvanHeLL/LSD-SLAM.git}.
\end{abstract}

\tableofcontents
\addcontentsline{toc}{section}{Abstract}


\newpage

\section{Introduction}

Simultaneous localization and mapping (SLAM) is one of the hardest and most important fields in computer vision. As pointed out by Montemerlo \cite{montemerlo2002fastslam}, SLAM is a fundamental problem of autonomous robots and therefore it plays a key role in robotics aswell as in augmented reality.

The challenge here is to aswell localize a robot within an unknown area of unknown scale and compute an internal global environment map for tracking. With computer vision, it is possible to solve that problem by creating a map of the environment in form of a 3D model. The camera images of the moving robot are used to observe the environment containing all the visible objects and structures. The creation of the map has to be done online, since it is neccessary to simultaneous localize the robot and update the global 3D map, frame by frame. The whole algorithm has a high computational complexity which makes it very challenging to achieve a suffienctly high framerate for an accurate tracking.

In Augmented Reality (AR), registration is one the most essential components. Since we were using the Oculus Rift as an Head-Mounted-Display (HMD), we had to capture the real world using undistorted camera images and render both camera streams onto the intern HMD screen. To achieve a stereoscopic view, we have used two fish-eye cameras IDS uEye UI-122-1LE-C-HQ. Both of them are positioned on the front plate of the Rift. Using this Virtual Reality (VR) device, we could  achieve a full-view AR immersion, since we are able to blend holograms everywhere at the intern screen with the real world. The development of our previous AR system is documented in our previous paper \cite{holl2016augmented}.

Microsoft's new  promising HoloLens is based on similar principles. It renders virtual objects onto the screen located within the lenses and tracks the 3D position in world space using the mounted camera. However, since the lenses of the HoloLens are transparent, there is no need for a digital camera stream. On the downside, the HoloLens can augment only a small area of the real world field-of-view (FOV), located in the center of the screen.

In section \ref{sec:theory}, we talk about Image Registration since this is a fundamental knowledge of SLAM systems. For illustration, we show some examples from the field of Medical Image Analysis. Then we cover SLAM methods in general and go into more detail about the monocular LSD-SLAM.

Section three \ref{sec:implementation} shows some implementation details about the integration and modifications, which have been done.

In section \ref{sec:conclusion} we summarize the experiences made during development and show some results. In section \ref{sec:future}, we list some possible future work.

 \subsection{Building upon our AR Oculus Rift System}

Considering our previous basic AR system \cite{holl2016augmented} with a modified Oculus Rift, and our own DirectX stereo engine, we were able to render and animate 3D graphic models super-imposed to the images captured by the cameras.

\begin{figure}[H]
  \centering
  \includegraphics[width=0.7\textwidth]{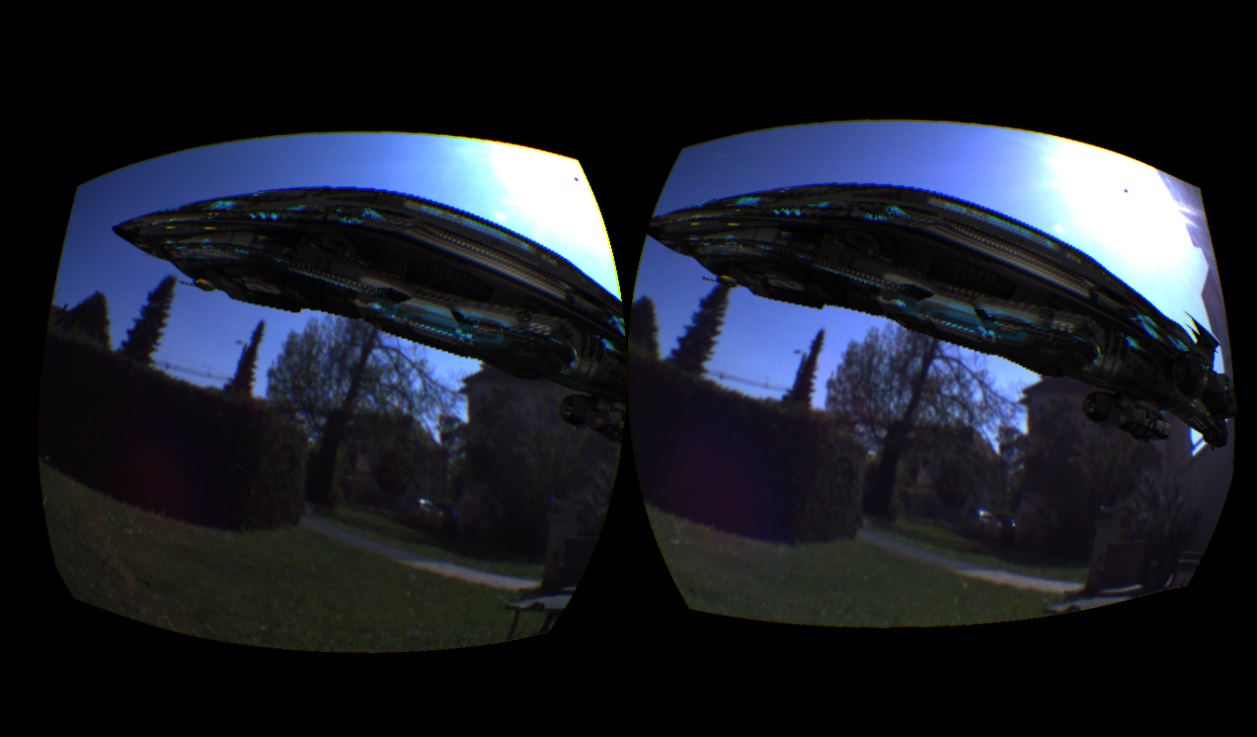}
  \caption{Results from our previously built AR system ''Augmented Reality Oculus Rift''. Image taken from \cite{holl2016augmented}.  }
  \label{fig:fish-eye}
\end{figure}

Using the Oculus Rift's internal gyro sensor, the augmentations could move consistently with the orientation of the user's head. But what makes AR systems really interesting, is a complete augmentation of the holograms into realworld space. To give a good illusion of augmentation, it is neccessary to consider translation motions as well.
 \begin{figure}[H]
    \centering
    \label{fig:hardware}
    \begin{subfigure}{0.45\textwidth}
        \includegraphics[width=\textwidth]{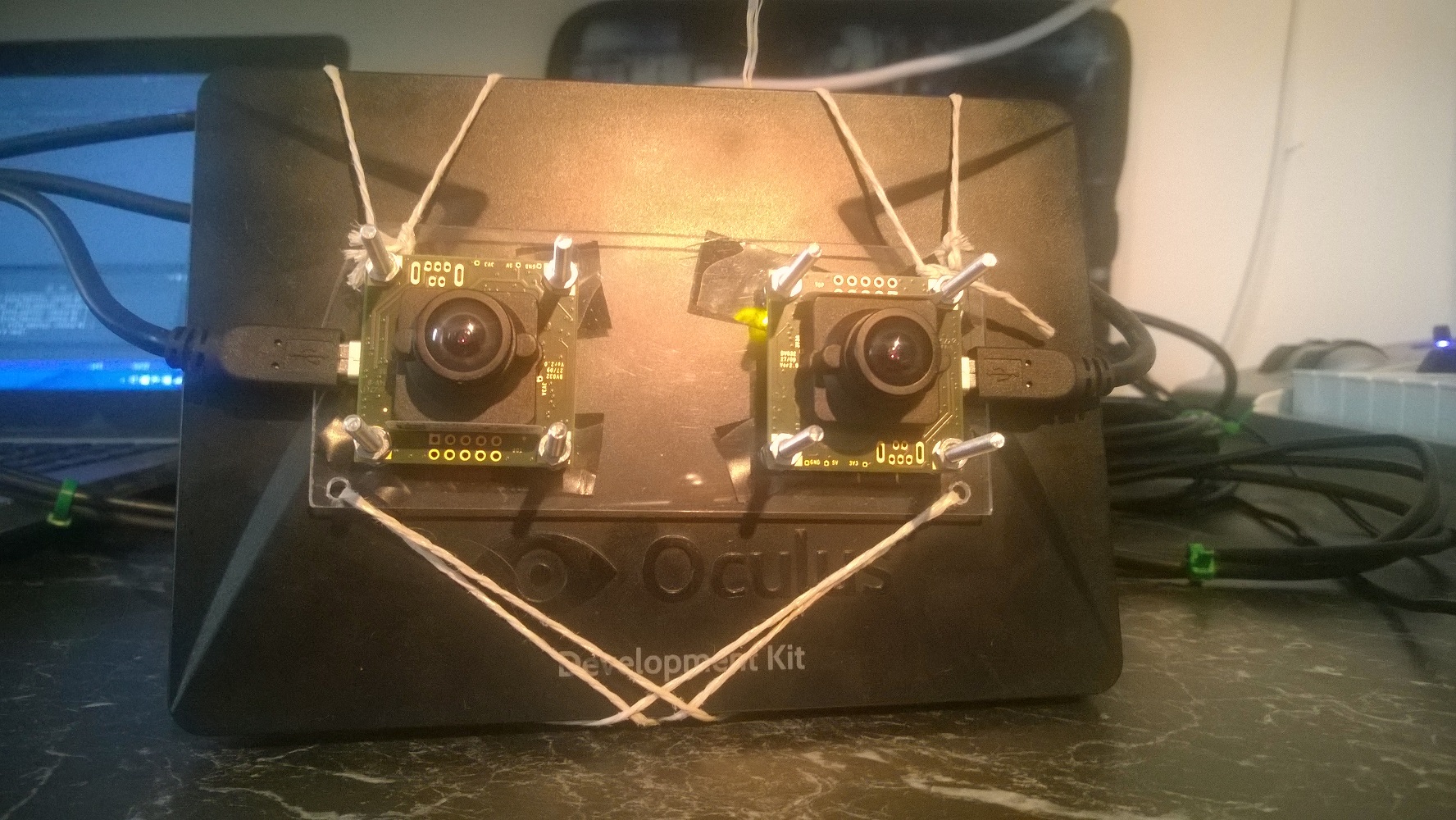}
        \label{fig:hardware1}
    \end{subfigure}
    ~ 
    \begin{subfigure}{0.45\textwidth}
        \includegraphics[width=\textwidth]{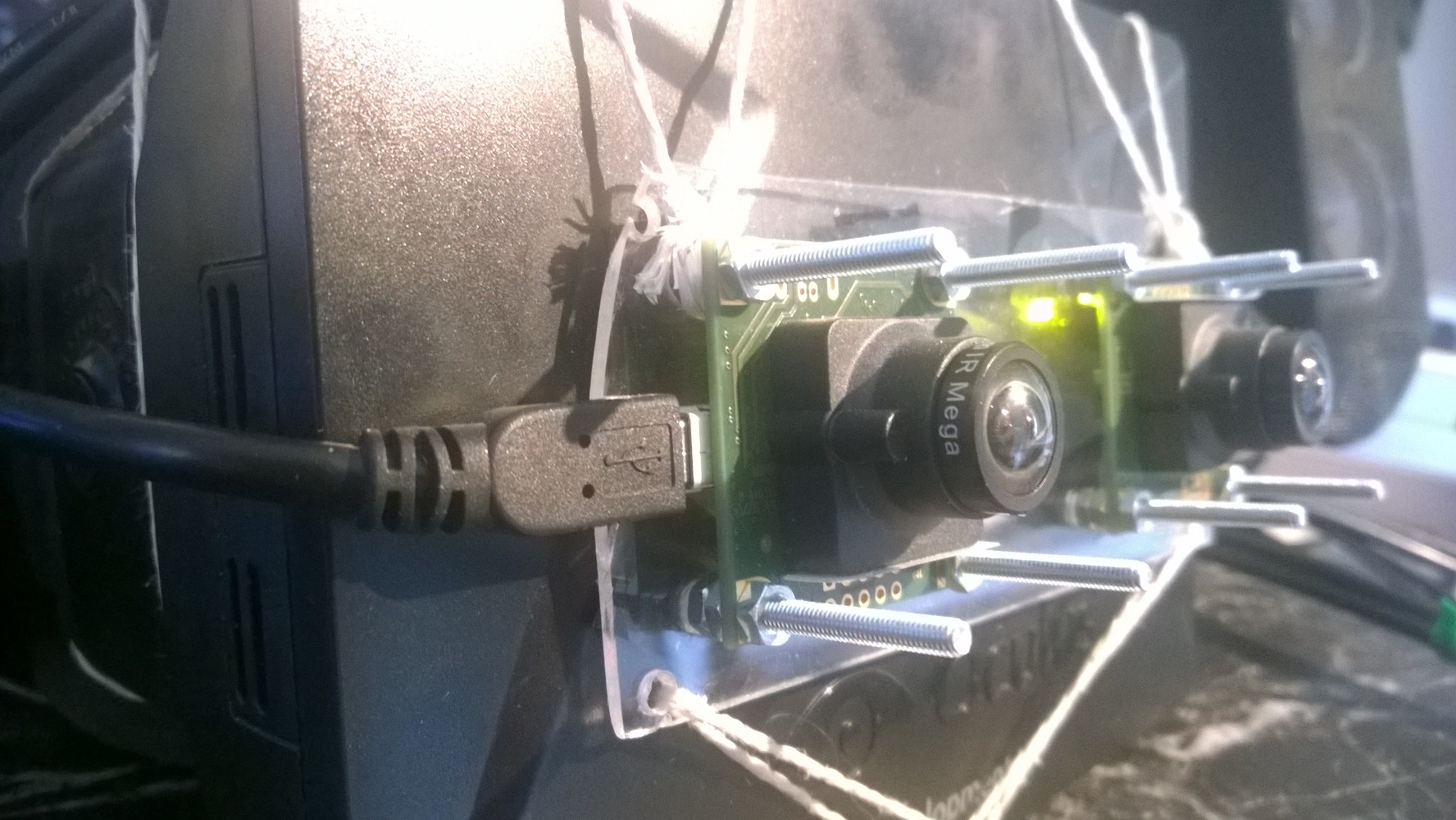}
        \label{fig:hardware2}
    \end{subfigure}
    \caption{Hardware Modifications on Oculus Rift DK1 for  ''Augmented Reality Oculus Rift''. Image property of Markus H\"oll }
\end{figure}
\label{fig:hardware}


\section{Theory}\label{sec:theory}

 \subsection{Vision in Augmented Reality} \label{sec:virtual-world}
 
Building  large-scale, accurate, responsive, and robust AR systems in real-time is challenging. What makes AR systems complex in computation is the real-world camera tracking, especially in large-scale areas without having information about the environment.

According to \cite{klein2007parallel}, the majority of AR systems (in 2007) operated with a prior knowledge of the environment the camera is located in. One possibility is to place a marker, maybe a card or a chessboard into the scene with known position like shown in \cite{kato1999marker}. By observing the marker, its prior known 3D world position can be used to measure the location of the observing camera in relation to the object in world space. The huge downside in approaches like that is, it is only possible to yield location and rotation of the camera while measuring the marker. That means, augmentation is not possible when the camera is currently not observing and measuring the object with prior known world coordinates. 

\begin{figure}[htpb]
  \centering
  \includegraphics[width=0.35\textwidth]{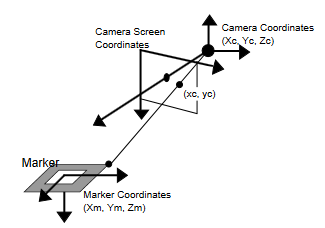}
  \caption{Relationship between marker coordinates and estimated camera coordinates. Image taken from \cite{kato1999marker}.  }
  \label{fig:fish-eye}
\end{figure}

Compared to these marker-based approaches, we wanted to integrate an extensible tracking, without having any prior information about the environment to fully augment reality. This need of extensible tracking opens the door to SLAM systems. 

How do we know about the 3D pose of the Rift in an entirely prior unknown environment, frame-by-frame in real-time? The answer is an algorithm called Monocular LSD-SLAM, introduced by Engel et al \cite{engel2014lsd}. It covers the process of recovering geometrical information from images. The camera could also be mounted on a robot and used for autonomous navigation. As an input sensor, we use the camera images. The underlying process of relating images to each other in LSD-SLAM is called Image Registration and is explained further in the next section with some illustrative examples from Medical Image Analysis.

 \subsection{Image Registration}
 
The term Image Registration defines the process of determining a one-to-one mapping or transformation between the coordinates in one space and those in another, such that points in the two spaces that correspond to the same point are mapped to each other. It is about to find a function, aligning a moving image with a second fixed image, such that a defined similarity measure is maximized. This underlying process can also be seen as part of LSD-SLAM, it registers images to each other to retrieve information about similarity transform \textbf{sim(3)} of our observed camera keyframes, achieved by image-aligning. 

 \begin{figure}[H]
  \centering
  \includegraphics[width=0.35\textwidth]{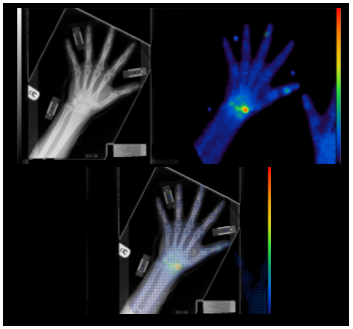}
  \caption{\textbf{Nuclear Medicine}: X-ray co-registered with corresponding bone scan. Image taken from \cite{hill2001medical}.  }
  \label{fig:medical_features}
\end{figure}
 
In this section, we talk shortly about the principal methods of Image Registration. We lean here mainly on the book Medical Image Registration from Hill \cite{hill2001medical}. In Medical Image Analysis, images from anatomical changing patients or different sensors  need to be registered to achieve Image Fusion and / or improve Image Guided Surgery based on AR systems.

 \begin{figure}[H]
  \centering
  \includegraphics[width=0.6\textwidth]{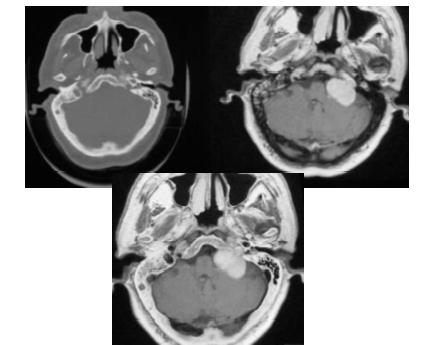}
  \caption{\textbf{Bottom}: Slice of Image Fusion from CT and MR Volumes  \textbf{Top Left}: CT \textbf{ TopRight}: MR. Image taken from \cite{hill2001medical}.  }
  \label{fig:medical_features}
\end{figure}

Image Registration can be done by four differnt methods. Three of them belong to rigid registration which are shortly explained here. Since nonlinear (nonrigid) registrations are not as relevant in our case, we skip them for the sake of this paper. The complexity of the methods increase in descending listing order.

 \subsubsection{Feature Based Rigid Registration}
 
 Placing anatomic or geometric landmarks on the human body makes it easy to identify these features (fiducials) in different images. Since the markers are very clearly to identify on every image, even if the images come from different sensors (CT-MR), the algorithm is straight forward by solving the Procrustes Problem with Point-Based Registration. Given two sets of \textbf{N} corresponding points \begin{math}\mathbf{P}\end{math} and \begin{math}\mathbf{Q}\end{math}, find the similarity transformation (scale factor \textbf{s}, rotation matrix \begin{math}\mathbf{R}\end{math} and translation vector \begin{math}\mathbf{P}\end{math}) that minimizes the mean squared distance between the points:
 
  \begin{equation}
    E_{Procrustes} =  \frac{1} {N_p}\sum_{i=1}^{N} \|\mathbf{sRp_i + t - q_i}\|_2^2
    \label{eq:surface_based}
  \end{equation}

 \begin{figure}[H]
  \centering
  \includegraphics[width=0.7\textwidth]{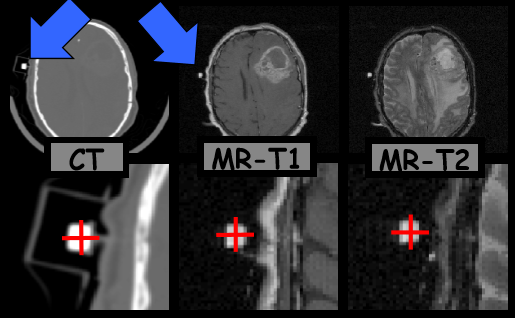}
  \caption{Identify corresponding feature points (placed landmarkers on the human head). Image taken from \cite{hill2001medical}.  }
  \label{fig:medical_features}
\end{figure}
 
\subsubsection{Surface Based Rigid Registration}

A good gemoetrical feature of an object is represented by its boundary. This approach can align large number of points in images but segmentation is necessary for surface extraction. The extracted surface is represented either as a point set, triangle set or as level set. Given a set of \begin{math}\mathbf{N_p}\end{math} surface points \begin{math}\mathbf{ \{p_i\} }\end{math} and a surface \textbf{Q}, it is about to find the rigid-body transformation \textbf{T}, where \textbf{T} is rotation matrix \textbf{R} and translation vector \textbf{t} that minimizes the mean squared distance between the points and the surface:

 \begin{equation}
    d(T) =  \frac{1} {N_p}\sum_{i=1}^{N_p} \|T(\mathbf{p}_i) -\mathbf{ q}_i\|_2^2
    \label{eq:surface_based}
  \end{equation} 
   \begin{equation}
    \mathbf{q_i} =  C(T(\mathbf{p}_i),Q)  ...  C = correspondence function
    \label{eq:c_surface_based}
  \end{equation}

  Iterative Closest Point algorithm yields the solution for this problem.
  
 \begin{figure}[H]
  \centering
  \includegraphics[width=0.9\textwidth]{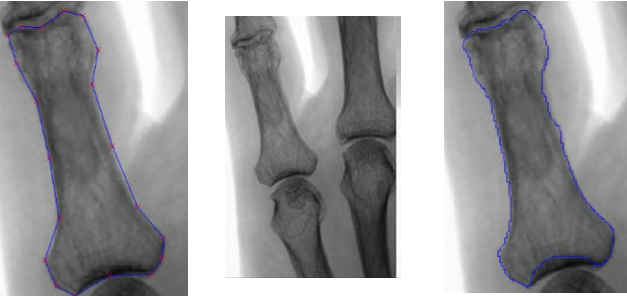}
  \caption{Surface \textbf{P}: data shape (contour) and Surface \textbf{Q}: model shape (high-resolution scan). Image taken from \cite{hill2001medical}.  }
  \label{fig:medical_surface}
\end{figure}
  
\subsubsection{Intensity Based Rigid Registration (Voxel Similarity)}

Intensity Based approaches are more complex, but also more powerful since more information is used. Voxel Similarity measures must distinguish between Single-modality registration and Multi-modality registration. The terms defines if the input images come from the same sensor, or different sensors. Depending on the domain, it is important to use suitable similarity assumptions. When speaking of LSD-SLAM, the scene consists of constantly changing intensities, but the sensor hardware (camera) stays the same.

   \begin{itemize}
      \item   \textbf{Single-Modality} means images are retrieved from the same sensor, e.g. CT-CT, MR-MR, PET-PET. etc.
            The main approach is to search a transformation \textbf{T}, which is determined by iteratively minimizing a voxel-based dissimilarity measure \textbf{C}. With Single-Modal, two different similarity assumptions about the intensity can be made, ''Identity'' and ''Linear'' which form the similarity metric.
            
 \textbf{ Identity} as a similarity assumption means, that the image intensity only differs by a Gaussian noise. The similarity metric of the Sum Of Squared Differences (SSD) can be used.
  \begin{equation}
    C(x; T) =  \frac{1} {N_p}\sum_{voxel_i}(I_A(x_i) - I_B(T(x_i)))^2
    \label{eq:ssd}
  \end{equation} 
            
      Though, it is sensitive to outliers, where the Sum Of Absolute Differences (SAD) is better suited.
      
  \textbf{ Linear} assumes a linear relationship between image intensities. As a similarity metric, the Normalized Cross Correlction (CC) is used.
  \begin{equation}
    CC(x; T) =  \frac{ \sum  \lbrack (I_A(x) -  \mu_A)(I_B(T(x)) - \mu_B) \rbrack }  { \sqrt{\sum (I_A(x) -  \mu_A)^2 \sum(I_B(T(x)) - \mu_B)^2} }
    \label{eq:cc}
  \end{equation} 
   
      \item   \textbf{Multi-Modality} (MR-CT, MR-PET, CT-PET, etc) assumes that image intensities are related by some unknown function or statistical relationship which is not known a-priori. Images are considered as  a (pixel-wise) probability distributions which can be estimated using a histogram. Knowing this, it is possible to create 2D Joint Histograms and retrieve knowledge about the Joint Probability p(a, b) of a pixel having intensity a in one image and intensity b in another image. 
      
       \begin{figure}[H]
  \centering
  \includegraphics[width=0.7\textwidth]{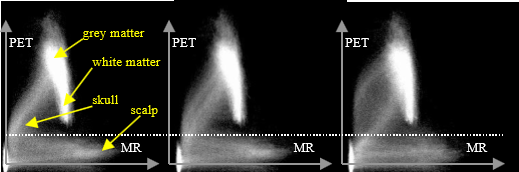}
  \caption{2D Joint Probability distribution of intensities for aligned MR and FDG PET volumes (\textbf{left}), misaligned with a 2mm translation (\textbf{middle}) and misaligned with a 5mm translation (\textbf{right}). Image taken from \cite{hill2001medical}.  }
  \label{fig:medical_surface}
\end{figure}
      
The amount of information in combined images A and B is described by the Joint Entropy. Registration is achieved by minimizing the joint entropy between both images. However, it is highly sensitive to the overlap of two images.

  \begin{equation}
    H(A, B) =  -\sum_{a} \sum_{b} p(a, b) log_2 p(a, b) 
    \label{eq:cc}
  \end{equation} 
  
  Mutual Information on the other hand, describes how well an image can be explained by another image and can be expressed in terms of marginal and joint probability distributions:
  
    \begin{equation}
    I(A, B) =  \sum_{a} \sum_{b} p(a, b) log_2 \frac{p(a, b) }  {p(a) p(b)} 
    \label{eq:cc}
  \end{equation} 
  
  Registration can be achieved by maximizing the Mutual Information between both images. Normalizing the Mutual Information leads to an independent term of this expression:
      
      \begin{equation}
    I(A, B) = \frac{H(A) + H(B) } {H(A, B)} 
    \label{eq:cc}
  \end{equation} 
  
  To optimize Voxel Similarity based techniques, Multi-Resolution optimazation leads to an acceleration.

      \end{itemize}

 \subsection{Simultaneous Localization and Mapping}

The task of localization solves the problem of estimating the 3D pose of a robot within an unkown environment. Mapping means computing a map of the environment in which the robot is located in. The map is then used as a reference for estimating 3D position. In principal, it is the same what humans do when trying to navigate within an unknown environment. Speaking of localization and mapping, means speaking of two different problems. However, as Davison \cite{davison2007monoslam} also pointed out, solving one of them requires solving both of them. Different SLAM approches are listet below.
 
\subsubsection{Feature-based monocular SLAM}
In feature-based monocular methods, the process of retrieving geometric information from images is splint into sequential tasks. The algorithm tries as a first step to extract feature observations from the input (image). From a set of keypoint obversations, it is then possible to estimate 3D pose of the current keypoint features and thus the camera world-coordinates by matching the features with the set of keypoint observations. This means, all the other image information except of the exctracted and observed features are thrown away, approaching however real-time performance since the complexity of the problem is reduced to the keypoint features. 

The successful project MonoSLAM from Davison \cite{davison2007monoslam} uses feature-based SLAM within a probabilistic framework running at a framerate of 30 Hz.

\begin{figure}[H]
  \centering
  \includegraphics[width=0.7\textwidth]{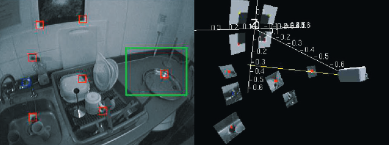}
  \caption{Detected feature patches of interest in camera stream (\textbf{left}) and visualized in world space coordinate system (\textbf{right}). Image taken from \cite{davison2007monoslam}.  }
  \label{fig:fish-eye}
\end{figure}

However, Engel \cite{engel2014lsd} clearly states out the big drawback on this method. Only features who match the defined feature type can be extracted (corners, line segments).

\subsubsection{Dense monocular SLAM (Direct)}

This intensity-based approach improves over the feature-based SLAM method by using all the information available in the image. It uses some measure derived directly from the intensity of the image pixels which leads to more information about the geometry. New image frames are tracked using whole-image alignment. There is no need for feature extraction as in the previous method. Due to the dense information, tracking accuracy and robustness is widely improved over feature-based methods. That makes direct methods very valuable for robotics or augmented reality systems.

The downside is the costly computation. As a comparison, the dense depthmap is similar to the output of a RGB-D camera shown in the following figure.

\begin{figure}[htpb]
  \centering
  \includegraphics[width=0.90\textwidth]{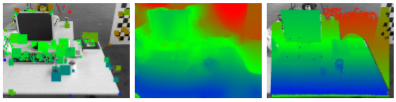}
  \caption{\textbf{Left}: keypoint depthmap \textbf{Center}: dense depthmap \textbf{Right}: RGB-D camera. Image taken from \cite{engel2013semi}.  }
  \label{fig:fish-eye}
\end{figure}

 \subsection{Monocular LSD-SLAM}

With Monocular Large-Scale Semi Dense SLAM, Engel  \cite{engel2014lsd, engel2013semi} proposed in 2013 the propably first feautreless real-time approach for monocular visual odometry, running with real-time framerates on a CPU which cuts out the need for high-parallel performance GPU hardware. The algorithm introduces 2 additional fundamental aspects. One is the estimation of semi-dense depthmaps which improves performance sigificantly over dense depthmaps. The second characteristic is then given by the large scale capability of the algorithm. 

The algorithm generates an intern global map, frame by frame, using direct image alignment to estimate the camera's 3D world position. This global map consists of keyframes as vertices with 3D similarity transform \textbf{sim(3)}, represented as a pose graph. For each localization estimation, the most up-to-date depthmap from the camera's recent keyframe is used as a reference for aligning the new captured image after motion movement. 

\begin{figure}[htpb]
  \centering
  \includegraphics[width=0.90\textwidth]{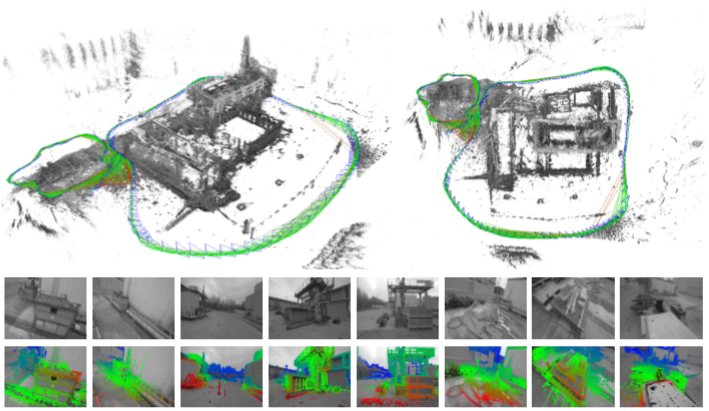}
  \caption{Camera trajectory with LSD-SLAM on medium scale, showing some estimated semi-dense keyframe depthmaps. Image taken from  \cite{engel2014lsd}.  }
  \label{fig:fish-eye}
\end{figure}

 \subsubsection{Semi-Dense}

The key idea behind this approach is not to build a dense depthmap using all intensities within the images, but to estimate a semi-dense inverse depth map for the current observed keyframe. The semi-dense inverse depthmap is modeled by using a Gaussian probability distribution which models one inverse depth hypothesis per pixel. The depthmap is estimated each frame and updated with a variable-baseline stere comparison. 

\begin{figure}[htpb]
  \centering
  \includegraphics[width=0.80\textwidth]{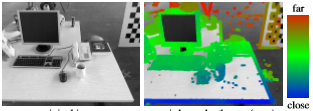}
  \caption{\textbf{Left}: original captured camera image. \textbf{Right}: semi-dense depth map (LSD-SLAM). Image taken from  \cite{engel2013semi}.  }
  \label{fig:fish-eye}
\end{figure}

\subsubsection{Large-Scale}

 This characteristic of Monocular LSD-SLAM solves one of the biggest challenges in SLAM because world scale can not be observed and it changes during runtime. As depth or stereo cameras are very limited in scale, to provide realible measurements, LSD-SLAM enables a smooth switch between different scales within the world. The scale problem is solved by using inherent correlation between scene depth and tracking accuracy. Keyframes have to be scaled such that the mean inverse depth is one and edges between keyframes are estimated as elements of \textbf{sim(3)} allowing to detect scale-drift. 
 
 \begin{figure}[htpb]
  \centering
  \includegraphics[width=0.9\textwidth]{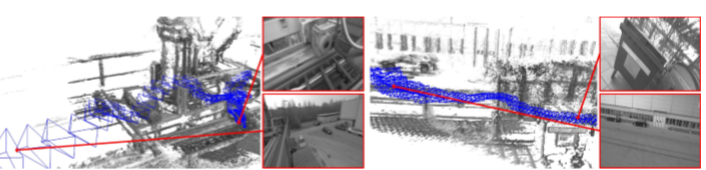}
  \caption{Two different scenes showing scale-differences. The keyframe's camera viewport is displayed in red rectangles.  Image taken from  \cite{engel2014lsd}.  }
  \label{fig:fish-eye}
\end{figure}

 \subsubsection{Complete Algorithm}

\begin{figure}[H]
  \centering
  \includegraphics[width=1.0\textwidth]{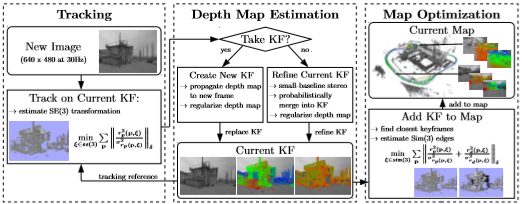}
  \caption{Illustration of the complete LSD-SLAM algorithm.  Image taken from  \cite{engel2014lsd}.  }
  \label{fig:complete_algorithm}
\end{figure}

To give an overview about the LSD-SLAM algorithm from Engel \cite{engel2014lsd}, we summarize the components shortly. The algorithm has 3 main components, \textbf{tracking}, \textbf{depth map estimation} and \textbf{map optimization}, illustrated in figure \ref{fig:complete_algorithm}

  \begin{itemize}
      \item \textbf{Tracking} executes the task of permanently tracking new camera images. The rigid body pose se(3) of each current keyframe is estimated by using the pose of the previous frame as a reference.
    \end{itemize}
      \begin{itemize}
         \item \textbf{Depth map estimation} computes the semi-dense depth of each current keyframe by filtering over many per-pixel, small-baseline stereo comparisons. It uses tracked frames, and refines the current keyframe. In the case the camera's motion was too large by moving too far, a new keyframe is initialized which replaces the old keyframe. The replacement is done by projecting points from existing, close-by keyframes onto the new keyframe.
      \end{itemize}
        \begin{itemize}
      \item \textbf{Map optimization} triggers when a new keyframe is replaced instead of depth refinement. It's purpose is to update the global map with the new tracked keyframe. Loop closures and scale-drift are detected by estimating a similarity transform sim(3) to close-by existing keyframes with scale-aware, direct sim(3)-image alignment.
    \end{itemize}

\lstset{language=C++,
       numbers=left,
        basicstyle=\ttfamily,
        keywordstyle=\color{blue}\ttfamily,
        commentstyle=\color{green}\ttfamily,
        frame=single,
        rulecolor=\color{black},    
        breaklines=true,
}

\section{Implementation}\label{sec:implementation}

The project is now split into 2 Visual Studio projects, each accessible on Github. The basic AR engine with Oculus Rift support  can be accessed via \url{https://github.com/MaXvanHeLL/ARift.git} and the modified LSD-SLAM project for the integration is uploaded here \url{https://github.com/MaXvanHeLL/LSD-SLAM.git}.

\subsection{Calibration}

Here we recap shortly the camera calibration process, documented in our previous paper \cite{holl2016augmented}. We have used the omnidirectional camera model from Scaramuzza \cite{scaramuzza2006toolbox} and have written an undistortion shader in HLSL for our engine. Thus, we could bypass the undistortion step for the LSD-SLAM, and benefit from a performance boost, since the undistortion shader runs highly parallel on the GPU.  We have computed the calibration with a polynomial function of fifth order. Distortion coefficients are already integrated in the projection function F.

  \begin{equation}
   F_{projection} =  a_0 + a_1p + a_2p^2 + a_3p^3 + a_4p^4 + a_5p^5
    \label{eq:omnidirectional}
  \end{equation} 
  
If the undistortion, integrated in the LSD-SLAM project, is desired, the camera matrix has to be built  with focal lengths and optical centers measured in pixels.

\[
Camera Matrix_{pinhole} =
 \begin{bmatrix}
    f_x & 0 & c_x  \\
    0 & f_y & c_y   \\
    0  & 0 & 1 \\
\end{bmatrix}
\]

Also the distortion coefficients are neccessary.

  \begin{equation}
   Distortion_{coefficients} =  (k_1,  k_2,  p_1,  p_2,  k_3)
    \label{eq:distortion_coefficients}
  \end{equation} 

Both, the camera matrix and the distortion coefficients are held in the file \textbf{out\_camera\_data.xml}, located in the data directory of the project root.

\subsection{Engine Integration}

In our previous paper \cite{holl2016augmented} we documented our complete AR stereo engine so far. In this section, we give an overview of the major new components due to the LSD-SLAM integration.

\begin{figure}[H]
  \centering
  \includegraphics[width=1.0\textwidth]{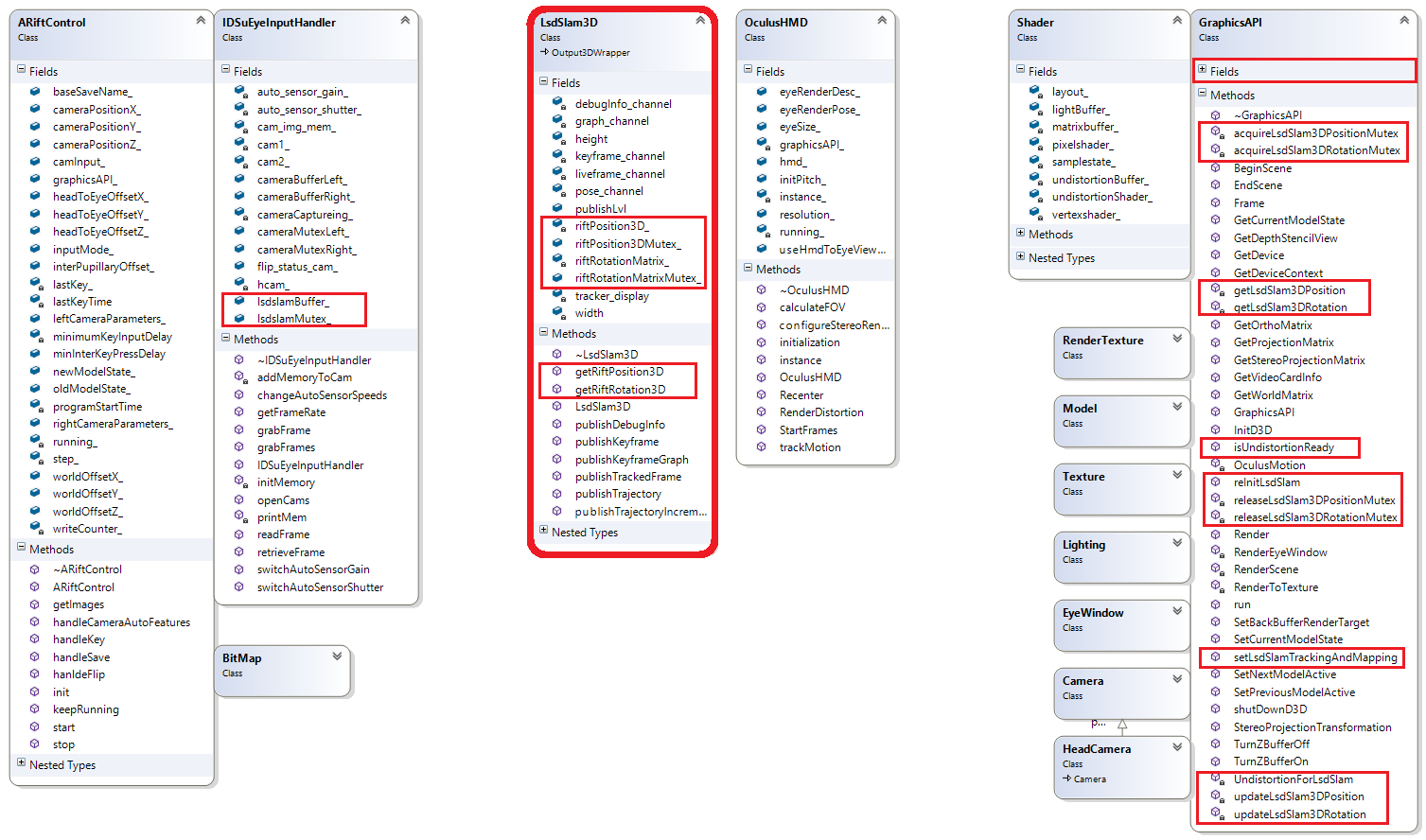}
  \caption{LSD-SLAM code integration. Image property of Markus H\"oll.  }
  \label{fig:program}
\end{figure}

The core of the integration is represented by the new class \textbf{LsdSlam3D}, illustrated in figure \ref{fig:program}.  It can be seen as an API to the dedicated Visual Studio project of LSD-SLAM and holds for us the important information about the camera's real world position and rotation.

Since the GraphicsAPI is constantly synchronizing the virtual camera with the now fully tracked LSD-SLAM camera, we added a mutex to ensure stable synchronization and guarantee always clean and valid data information.

Additionally, we have added another mutex, used for the camera images. Since the camera stream is running in another thread, we also synchronized the data exchange between the image capturing and SLAM operation.

\subsection{Initialization} \label{sec:init}

In the main function we set up all the needed LSD-SLAM properties. First of all, the \textbf{CvCapture} property, which holds information about the tracked monocular camera. The  camera is referenced wit an ID, starting from 0 to ascending order. Also the capture frame dimensions are set here with 640 x 480.

After that, we initialize the camera stream thread, used by LSD-SLAM. Since we modified the original lsd\_slam::OpenCVImageStreamThread, we introduced a new one called \textbf{lsd\_slam::IDSuEyeCameraStreamThread}. The slight modifications to our custom input stream are documented in section \ref{sec:codemodifications}.

Now we are about to set the monocular camera buffer to the IDSuEyeCameraStreamThread. The camera stream thread will from now on constantly read out the frames stored in the undistortion buffer, retrieved from the HLSL shader.

To complete the initalization, we allocate our new introduced \textbf{LsdSlam3D} API object where we will retrieve the desired 3D camera translation and rotation as an output.

\begin{lstlisting}
int main(int, char**)
{
  // init DirectX Graphics, Cameras and OculusHMD 
  [...]
  
  CvCapture* capture = cvCaptureFromCAM(LsdSlam_CAM)
  cvSetCaptureProperty(capture, CV_CAP_PROP_FRAME_WIDTH, 640);
  cvSetCaptureProperty(capture, CV_CAP_PROP_FRAME_HEIGHT, 480);

  lsd_slam::IDSuEyeCameraStreamThread* inputStream = new lsd_slam::IDSuEyeCameraStreamThread();
  inputStream->setCameraCapture(capture);
  if (LsdSlam_UNDISTORTION) // use undistorted imgs
  	inputStream->setIDSuEyeCameraStream(undistBuffer_, undistBufferMutex_, true);
  else // use raw images
  	inputStream->setIDSuEyeCameraStream(cont.camInput_->lsdslamBuffer_, cont.camInput_->lsdslamMutex_, false);
  inputStream->run();

  // Init LSD-SLAM with outputstream and inputstream
  lsd_slam::Output3DWrapper* outputStream = new lsd_slam::LsdSlam3D(inputStream->width(), inputStream->height());
  lsd_slam::LiveSLAMWrapper slamNode(inputStream, outputStream);
  dx11->setLsdSlamTrackingAndMapping(outputStream);	
}
\end{lstlisting}

\subsection{Render Loop Update}\label{sec:stereorenderloop}

Some program controls allow us to switch between LSD-SLAM rotation and gyro sensor rotation from the Oculus Rift, aswell as to decide if we want to use LSD-SLAM translation or only the virtual head camera. Since we can set those controls individually, the \textbf{Frame()} function has been modified accordingly:

\begin{lstlisting}
bool GraphicsAPI::Frame()
{
  // [ Lsd-Slam ] Virtual Camera Update ---------------------------------
  if (LsdSlam_Rotation)
    updateLsdSlam3DRotation();
  else
    OculusMotion();

  // [ Lsd-Slam ] Virtual Camera Update ---------------------------------
  if (LsdSlam_Translation)
    updateLsdSlam3DPosition();
  else
    headCamera_->SetPosition(ariftcontrol_->cameraPositionX_, ariftcontrol_->cameraPositionY_, ariftcontrol_->cameraPositionZ_);

  // Render the virtual scene.
  Render();
}
\end{lstlisting}

As the name suggests, the \textbf{updateLsdSlam3DRotation()} method updates internally the virtual camera rotation matrix with the one retrieved from LSD-SLAM. Also we convert the \textbf{Eigen::MatrixXd*} to our matching coordinate system convention.

\begin{lstlisting}
void GraphicsAPI::updateLsdSlam3DRotation()
{
  // waits internally or throws error code
  if (!acquireLsdSlam3DRotationMutex())
    return;

  XMFLOAT3X3 cam.Rot.Matrix = {};
  if (lsdslam3D_)
  {
    Eigen::MatrixXd* riftRot.Matrix = dynamic_cast<lsd_slam::LsdSlam3D*>(lsdslam3D_)->getRiftRotation3D();
    cam.Rot.Matrix._11 = (*riftRot.Matrix)(0,0);
    cam.Rot.Matrix._12 = (*riftRot.Matrix)(1,0);
    cam.Rot.Matrix._13 = (*riftRot.Matrix)(2,0);
    cam.Rot.Matrix._21 = (*riftRot.Matrix)(0,1);
    cam.Rot.Matrix._22 = (*riftRot.Matrix)(1,1);
    cam.Rot.Matrix._23 = (*riftRot.Matrix)(2,1);
    cam.Rot.Matrix._31 = (*riftRot.Matrix)(0,2);
    cam.Rot.Matrix._32 = (*riftRot.Matrix)(1,2);
    cam.Rot.Matrix._33 = (*riftRot.Matrix)(2,2);
  		
    cam.Rot.Matrix._32 *= (-1);
    cam.Rot.Matrix._23 *= (-1);
    cam.Rot.Matrix._12 *= (-1);
    cam.Rot.Matrix._21 *= (-1);
  }
  
  headCamera_->LsdSlamRotationMatrix_ = rotationMatrix;
   
   releaseLsdSlam3DRotationMutex();
}
\end{lstlisting}

What happens in \textbf{updateLsdSlam3DTranslation()} is basically a smooth 3D movement implementation and a reference comparison with LSD-SLAM data. We store a \textbf{lsdslam\_reference} which is directly compared with the accurat retrieved \textbf{riftPosition} from LSD-SLAM . Since we wanted to achieve a smooth movement  we have added some thresholding value to erase small correction errors. Further we invented a scaling factor for the movement, to be not as limited and scale in different worlds accordingly. Of course, when we don't have a \textbf{lsdslam\_reference} yet, we are in the initialization phase and just store the current position as the origin reference.

\begin{lstlisting}
void GraphicsAPI::updateLsdSlam3DPosition()
{
  // waits internally or throws error code
  if (!acquireLsdSlam3DPositionMutex())
    return;
		
  bool pos_x_changed = false;
  bool pos_y_changed = false;
  bool pos_z_changed = false;

  // calling LsdSlam3D API internally
  lsd_slam::LsdSlam3D::RiftPosition3D* riftPosition = getLsdSlam3DPosition();
  if (lsdslam_init_ == true)
  {
    lsdslam_ref.x = riftPosition->x;
    lsdslam_ref.y = riftPosition->y;
    lsdslam_ref.z = riftPosition->z;
    lsdslam_init_ = false;
  }
	
  XMFLOAT3 camera_position = headCamera_->GetPosition();
  float move_direction = headCamera_->headToEyeOffset_.rotationY_ * 0.0174532925f;

  // [Move] Forward / Backward -------------------
  if ((riftPosition->z > (lsdslam_reference_.z + LsdSlam_Threshold))) // Move Forwards
  {
    float distance = riftPosition->z - lsdslam_reference_.z;
    float new_position_x = (camera_position.x) + sinf(move_direction) * (distance * LsdSlam_ScaleFactor);
    float new_position_z = (camera_position.z) + cosf(move_direction) * (distance * LsdSlam_ScaleFactor);
    headCamera_->SetPositionX((new_position_x));
    headCamera_->SetPositionZ((new_position_z));
    pos_x_changed = true;
    pos_z_changed = true;
  }
  else if ((riftPosition->z < (lsdslam_reference_.z - LsdSlam_Threshold))) // Move Backwards
  {
    float distance = lsdslam_reference_.z - riftPosition->z;
    float new_position_x = (camera_position.x) - sinf(move_direction) * (distance * LsdSlam_ScaleFactor);
    float new_position_z = (camera_position.z) - cosf(move_direction) * (distance * LsdSlam_ScaleFactor);
    headCamera_->SetPositionX((new_position_x));
    headCamera_->SetPositionZ((new_position_z));
    pos_x_changed = true;
    pos_z_changed = true;
  }
  // [Strafe] Left / Right -------------------
  if ((riftPosition->x > (lsdslam_reference_.x + LsdSlam_Threshold))) // Strafe Right
  {
    float distance = riftPosition->x - lsdslam_reference_.x;
    float new_position_x = (camera_position.x) + cosf(move_direction) * (distance * LsdSlam_ScaleFactor);
    float new_position_z = (camera_position.z) + sinf(move_direction) * (distance * LsdSlam_ScaleFactor);
    headCamera_->SetPositionX((new_position_x));
    if(!pos_z_changed)
      headCamera_->SetPositionZ((new_position_z));
    pos_x_changed = true;
    pos_z_changed = true;
  }
  else if ((riftPosition->x < (lsdslam_reference_.x - LsdSlam_Threshold))) // Strafe Left
  {
    float distance = lsdslam_reference_.x - riftPosition->x;
    float new_position_x = (camera_position.x) - cosf(move_direction) * (distance * LsdSlam_ScaleFactor);
    float new_position_z = (camera_position.z) - sinf(move_direction) * (distance * LsdSlam_ScaleFactor);
    headCamera_->SetPositionX((new_position_x));
    if (!pos_z_changed)
      headCamera_->SetPositionZ((new_position_z));
    pos_x_changed = true;
    pos_z_changed = true;
  }
  // [Fly] Up / Down -------------------
  if ((riftPosition->y > (lsdslam_reference_.y + LsdSlam_Threshold))) // Down
  {
    float distance = riftPosition->y - lsdslam_reference_.y;
    float new_position_y = (camera_position.y) -  (distance * LsdSlam_ScaleFactor);
    headCamera_->SetPositionY((new_position_y));
    pos_y_changed = true;
  }
  else if ((riftPosition->y < (lsdslam_reference_.y - LsdSlam_Threshold))) // Up
  {
    float distance = lsdslam_reference_.y - riftPosition->y;
    float new_position_y = (camera_position.y) + (distance * LsdSlam_ScaleFactor);
    headCamera_->SetPositionY((new_position_y));
    pos_y_changed = true;
  }
  
  // update LsdSlam Reference Position
  if (pos_x_changed)
    lsdslam_reference_.x = riftPosition->x;
  if (pos_y_changed)
    lsdslam_reference_.y = riftPosition->y;
  if (pos_z_changed)
    lsdslam_reference_.z = riftPosition->z;

  releaseLsdSlam3DPositionMutex();
}
\end{lstlisting}

\subsection{Undistortion ShaderBuffer for LSD-SLAM}\label{sec:undistortionbuffer}

Since we have used our own undistortion images retrieved from our fragment shader, we had to adapt the region-of-interest (ROI) of the undistorted images to ensure that we only pass fully valid image data to our \textbf{lsd\_slam::IDSuEyeCameraStreamThread}.

\begin{figure}[H]
  \centering
  \includegraphics[width=0.8\textwidth]{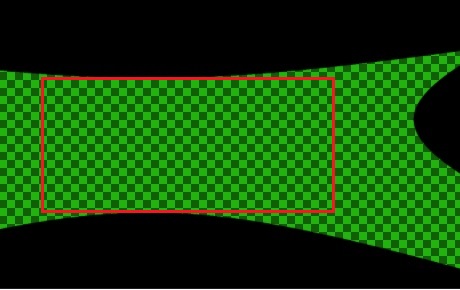}
  \caption{Undistortion mapping and ROI extraction. Image property of Markus H\"oll.  }
  \label{fig:undistortionmapping}
\end{figure}

After the undistortion, our image buffer itself holds of course black borders which correct the distortion, illustrated in figure \ref{fig:undistortionmapping}. With the parameter \textbf{camID} we can control which of our cameras we want to use for the LSD-SLAM, which changes also the ROI parameters due to undistortion.

The undistorted images were so far fully placed in the GPU memory, since they were computed from our fragment shader. Therefore, we had to setup an explicit mapped CPU access to copy data from the GPU memory.

After we have copied the image data to from the GPU, we converted the images' data buffer to an \textbf{IplImage* frame} where we then can extract the correct ROI, depending on which camera we use for tracking.

For the last step, we resized the image to 640 x 480 which are the suggested dimensions for LSD-SLAM and copy the images to the image buffer.

\begin{lstlisting}
void GraphicsAPI::UndistortionForLsdSlam(int camID)
{
  ID3D11Resource* renderBuffer;
  if (camID == 1)
    renderTextureLeft_->GetRenderTargetView()->GetResource(&renderBuffer);
  else
    renderTextureRight_->GetRenderTargetView()->GetResource(&renderBuffer);

  D3D11_TEXTURE2D_DESC texDesc;
  texDesc.ArraySize = 1;
  texDesc.BindFlags = 0;
  texDesc.CPUAccessFlags = D3D11_CPU_ACCESS_READ;
  texDesc.Format = DXGI_FORMAT_R8G8B8A8_UNORM;
  texDesc.Width = screenWidth_;
  texDesc.Height = screenHeight_;
  texDesc.MipLevels = 1;
  texDesc.MiscFlags = 0;
  texDesc.SampleDesc.Count = 1;
  texDesc.SampleDesc.Quality = 0;
  texDesc.Usage = D3D11_USAGE_STAGING;

  ID3D11Texture2D* undistortedShaderTex;
  device_->CreateTexture2D(&texDesc, 0, &undistortedShaderTex);
  devicecontext_->CopyResource(undistortedShaderTex, renderBuffer);

  // Map Resource From GPU to CPU
  D3D11_MAPPED_SUBRESOURCE mappedResource;
  if (FAILED(devicecontext_->Map(undistortedShaderTex, 0, D3D11_MAP_READ, 0, &mappedResource)))
    std::cout << "Error: [CAM 2] could not Map Rendered Camera ShaderResource for Undistortion" << std::endl;

  unsigned char* buffer = new unsigned char[(screenWidth_ * screenHeight_ * CAMERA_CHANNELS)];
  unsigned char* mappedData = reinterpret_cast<unsigned char*>(mappedResource.pData);
  memcpy(buffer, mappedData, (screenWidth_ * screenHeight_ * 4));
  devicecontext_->Unmap(undistortedShaderTex, 0);

  // OpenCV IplImage* Convertion
  IplImage* frame = cvCreateImageHeader(cvSize(screenWidth_, screenHeight_), IPL_DEPTH_8U, CAMERA_CHANNELS);
  frame->imageData = (char*)buffer;
  cvSetData(frame, buffer, frame->widthStep);
  cv::Mat mymat = cv::Mat(frame, true);

  // Extract ROI
  cv::Rect roi;
  switch (camID)
  {
    case 1: // Left Camera
    if (HMD_DISTORTION)
      roi = cv::Rect(300, 230, screenWidth_ - 300, screenHeight_ - 410);
    else
      roi = cv::Rect(10, 170, screenWidth_ - 10, screenHeight_ - 250);
    break;

    case 2: // Right Camera
      if (HMD_DISTORTION)
        roi = cv::Rect(5, 230, screenWidth_ - 170, screenHeight_ - 450);
      else
        roi = cv::Rect(5, 150, screenWidth_ - 10, screenHeight_ - 290);
      break;
  
    default:
      roi = cv::Rect(0, 0, screenWidth_, screenHeight_);
  }
  cv::Mat roi_mat = mymat(roi);

  // Resize for LSD-SLAM 640x480
  IplImage* resized_ipl = cvCreateImage(cvSize(640, 480), IPL_DEPTH_8U, CAMERA_CHANNELS);
  IplImage* ipl_roi = new IplImage(roi_mat);
  cvResize((IplImage*)ipl_roi, resized_ipl, CV_INTER_LINEAR);
  cv::Mat lsdslam_image = cv::Mat(resized_ipl, true);

  WaitForSingleObject(undistortedShaderMutex_, INFINITE);
  memcpy(undistortedShaderBuffer_, lsdslam_image.data, (640 * 480 * CAMERA_CHANNELS));
  ReleaseMutex(undistortedShaderMutex_);

  if (!undistortionReady_)
    undistortionReady_ = true;

  if (undistortedShaderTex)
    undistortedShaderTex->Release();

  if (buffer)
   delete(buffer);

  if (resized_ipl)
    cvReleaseImage(&resized_ipl);

  if (ipl_roi)
    delete(ipl_roi);
}

\end{lstlisting}

\subsection{LSD-SLAM Code Modifications} \label{sec:codemodifications}

Generally, the camera capturing and undistortion process is implemented in the \textbf{OpenCVImageStreamThread} but we already mentioned above that we have used our own undistortion, retrieved from our undistortion shader and supply LSD-SLAM with a buffer of those undistorted images. We show the code modifications within the LSD-SLAM project itself. For this purpose, we introduced a new class \textbf{lsd\_slam::IDSuEyeCameraStreamThread}.

\subsubsection{lsd\_slam::IDSuEyeCameraStreamThread}\label{sec:camerastreamthread}

The constructor makes it pretty clear which data is neccessary. The cameraBuffer\_ and cameraMutex\_ are pointers to instances where the thread should read data from. We just copy the data from cameraBuffer\_ to the internally used frameBuffer\_.

iDSuEyeFrame\_ is the final data type on which LSD-SLAM is processing on and resizedFrame\_ is needed in some cases when the image dimensions are not 640 x 480.

Finally, use\_shaderUndistortion\_ just tells us if we want to use the undistorted images supplied by our shader, or might want to use raw image data or something else for any reason.

\begin{lstlisting}
  IDSuEyeCameraStreamThread::IDSuEyeCameraStreamThread()
  {
    cameraBuffer_ = 0;
    cameraMutex_ = 0;
    frameBuffer_ = 0;
    iDSuEyeFrame_ = 0;
    resizedFrame_ = 0;
    use_shaderUndistortion_ = false;
  }
\end{lstlisting}

This method- \textbf{setIDSuEyeCameraStream()} - is called in the configuration phase directly from \textbf{main}, to set the corresponding pointers and configs for camera streaming.

\begin{lstlisting}
 void IDSuEyeCameraStreamThread::setIDSuEyeCameraStream(unsigned char* cameraBuffer, void* cameraMutex, bool use_shaderUndistortion)
{
  cameraBuffer_ = cameraBuffer;
  cameraMutex_ = cameraMutex;
  use_shaderUndistortion_ = use_shaderUndistortion;
  
  if (use_shaderUndistortion_)
    frameBuffer_ = new char[LsdSlam_BUFFER_LENGTH];
  else
  {
    frameBuffer_ = new char[CAMERA_BUFFER_LENGTH];
    resizedFrame_ = cvCreateImage(cvSize(LsdSlam_CAM_WIDTH, LsdSlam_CAM_HEIGHT), IPL_DEPTH_8U, CAMERA_CHANNELS);
  }
}
\end{lstlisting}

The \textbf{run()} method is calling the actual runtime loop of the thread.

\begin{lstlisting}
void IDSuEyeCameraStreamThread::run()
{
  boost::thread thread(boost::ref(*this));
 }
\end{lstlisting}

In the beginning before the \textbf{Loop} starts, we check if we have valid camera objects and jump then into the runtime loop, which overrides the actual image streaming to the LSD-SLAM. In case of using the undistortiion images from the shader, there is not much happening here since everything has been prepared previously from the \textbf{GraphicsAPI}.

However, if we might want to use the image raw data for tracking or something else, we have to sample the images to fit the suggested dimensions.

\begin{lstlisting}
void IDSuEyeCameraStreamThread::operator()()
{
  WaitForSingleObject(cameraMutex_, INFINITE);
  if (!cameraBuffer_ || !cameraMutex_)
  {
    assert("NO valid IDSuEye Camerabuffer found for Tracking and Mapping");
    return;
  }
  ReleaseMutex(cameraMutex_);
  while (1)
  {
    TimestampedMat bufferItem;
    bufferItem.timestamp = Timestamp::now();
			
    // Use Undistortion provided by our Fragment Shader ROI (640 x 480)
    if (use_shaderUndistortion_)
    {
      WaitForSingleObject(cameraMutex_, INFINITE);
      memcpy(frameBuffer_, cameraBuffer_, LsdSlam_BUFFER_LENGTH);
      ReleaseMutex(cameraMutex_);
      iDSuEyeFrame_ = cvCreateImageHeader(cvSize(LsdSlam_CAM_WIDTH, LsdSlam_CAM_HEIGHT), IPL_DEPTH_8U, CAMERA_CHANNELS);
      iDSuEyeFrame_->imageData = frameBuffer_;
      iDSuEyeFrame_->imageDataOrigin = iDSuEyeFrame_->imageData;
     }
    else // use raw image data from cams (752 x 480)
    {
      WaitForSingleObject(cameraMutex_, INFINITE);
      memcpy(frameBuffer_, cameraBuffer_, CAMERA_BUFFER_LENGTH);
      ReleaseMutex(cameraMutex_);
      iDSuEyeFrame_ = cvCreateImageHeader(cvSize(CAMERA_WIDTH, CAMERA_HEIGHT), IPL_DEPTH_8U, CAMERA_CHANNELS);
      iDSuEyeFrame_->imageData = frameBuffer_;
      iDSuEyeFrame_->imageDataOrigin = iDSuEyeFrame_->imageData;
      iDSuEyeFrame_ = ResizeFrame(iDSuEyeFrame_, LsdSlam_CAM_WIDTH, LsdSlam_CAM_HEIGHT);
    }
    bufferItem.data = iDSuEyeFrame_;
    imageBuffer->pushBack(bufferItem);
  }
  exit(0);
}
\end{lstlisting}

A short method for re-sampling an IplImage to the desired image dimensions. In our case, we would downsample from 752 x 480 to 640 x 480.

\begin{lstlisting}

IplImage* IDSuEyeCameraStreamThread::ResizeFrame(IplImage* src_frame, int new_width, int new_height)
{
  if (new_width != LsdSlam_CAM_WIDTH || new_height != LsdSlam_CAM_HEIGHT)
    resizedFrame_ = cvCreateImage(cvSize(new_width, new_height), src_frame->depth, src_frame->nChannels);
  cvResize(src_frame, resizedFrame_, CV_INTER_LINEAR);
  return resizedFrame_;
}
\end{lstlisting}

\subsection{Light Shading Model - Rendering Equation}\label{sec:lightshading}

As a very small gimmick, we also modified our fragment shader from the previously used Lambert Shading to a Phong Shading model. This reflection model can be implemented without any effort actually but gives quite a more realistic look to the holograms due to the reflection term which yields specular highlights. Since it is still a local illumination model, an approximation of global illumination in form of an ambient light term is added to the rendering equation. 

Generally, the shading model has 3 components, a diffuse illumination \textbf{\begin{math}\mathbf{k_d}\end{math}}, a specular reflection component \textbf{\begin{math}\mathbf{k_s}\end{math}} and an ambient term \textbf{\begin{math}\mathbf{k_a}\end{math}}. This gives a formal expression like \ref{eq:phong_simple}.

 \begin{equation}
    I_{ShadingModel} = I_{diffuse} + I_{ambient} + I_{specular}.
    \label{eq:phong_simple}
  \end{equation}

Formulating the model more precisely for each component, we get the render equation according to formula \ref{eq:phong_shading}. 

 \begin{equation}
    I_{PhongShading} =  k_aI_a + I_{direct}(K_d(\vec{L} \cdot \vec{N}) + k_s(\vec{V} \cdot \vec{R})^n)
    \label{eq:phong_shading}
  \end{equation} 
  
  \textbf{\begin{math}{ \vec{\mathbf{L}}}\end{math}} is a vector from the surface point to the emmitting light source. \textbf{\begin{math}{\vec{\mathbf{N}}}\end{math}} is the normal vector of the surface point. \textbf{\begin{math}{\vec{\mathbf{V}}}\end{math}} stands for a direction vector, pointing towards the observer (viewer) and \textbf{\begin{math}{\vec{\mathbf{R}}}\end{math}} is the direction of the reflected light which the light ray takes from this surface point on.

\subsection{New Engine Configurations}\label{sec:config}
A list of new configuration mechanics, added during the integration process:

   \begin{itemize}
      \item\texttt{LsdSlam\_Threshold} - configures a threshold which eliminates smaller correction errors in the tracking
      \item\texttt{LsdSlam\_ScaleFactor} - configure the scaling from the virtual movement, retrieved from LSD-SLAM
      \item\texttt{LsdSlam\_UNDISTORTION} - provide LSD-SLAM with undistorted shader images
      \item \texttt{LsdSlam\_Translation} - use translation from LSD-SLAM or not
      \item \texttt{LsdSlam\_Rotation} - use rotation from LSD-SLAM or Oculus Rift's gyro sensor
      \item\texttt{VSYNC\_ENABLED} - enables / disables vertical synchronization
   \end{itemize}

\subsection{New Libraries}\label{sec:library}

   \begin{itemize}
      \item LSD-SLAM
      \item OpenCV
      \item boost
      \item g2o
   \end{itemize}

\section{Conclusion}\label{sec:conclusion}

We are really satisfied with the results. The integration process actually worked quite good, however it was still some coding work to adapt the engine for the use of LSD-SLAM. That's mainly because we have used an omnidirectional camera model for our high degree fisheye-lenses to compute the calibration. And since we already computed the undistortion with a fragment shader in HLSL we wanted to use them for LSD-SLAM aswell. So basically, we had to tweak the image streaming to the algorithm and could not use the prepared OpenCVImageStreamThread.

The project was shown also at the \textbf{OpenLAB-Night 2016} of the \textbf{Institute for Computer Graphics and Computer Vision (ICG)} at \textbf{TU Graz}. The guests could try it out and we received alot of good feedback. The people showed alot of interest in the project and generally in AR itself.

Tracking the real-world camera and synchronizing it's 3D movement with the virtual head camera worked pretty well. It still feels exciting to walk towards 3D holograms which are augmented in the real world, going around them and crouch to see them from below.

The following figures show some final tracking results. The digimon model and texture file from the result images are taken from web source models-resource.com under educational usage purposes. The chessboard which sometimes is in the scene was only used for camera calibration.

 \begin{figure}[H]
    \centering
    \label{fig:near}
    \begin{subfigure}{0.4\textwidth}
        \includegraphics[width=\textwidth]{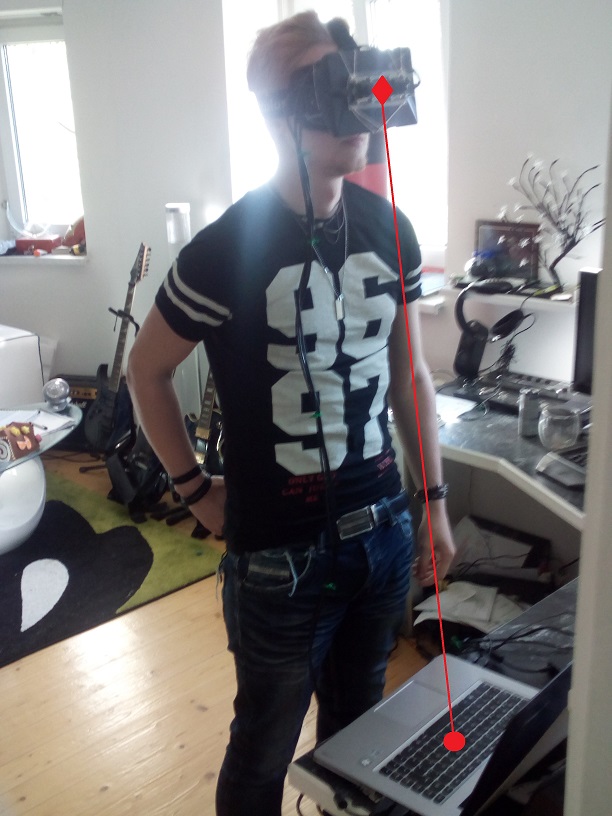}
        \label{fig:near1}
    \end{subfigure}
    ~ 
    \begin{subfigure}{0.55\textwidth}
        \includegraphics[width=\textwidth]{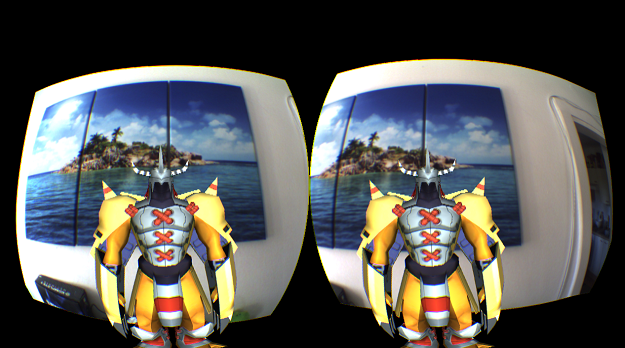}
        \label{fig:near2}
    \end{subfigure}
    \caption{Results: \textbf{near} distance. Image property of Markus H\"oll }
\end{figure}
\label{fig:near}

 \begin{figure}[H]
    \centering
    \label{fig:crouch}
    \begin{subfigure}{0.4\textwidth}
        \includegraphics[width=\textwidth]{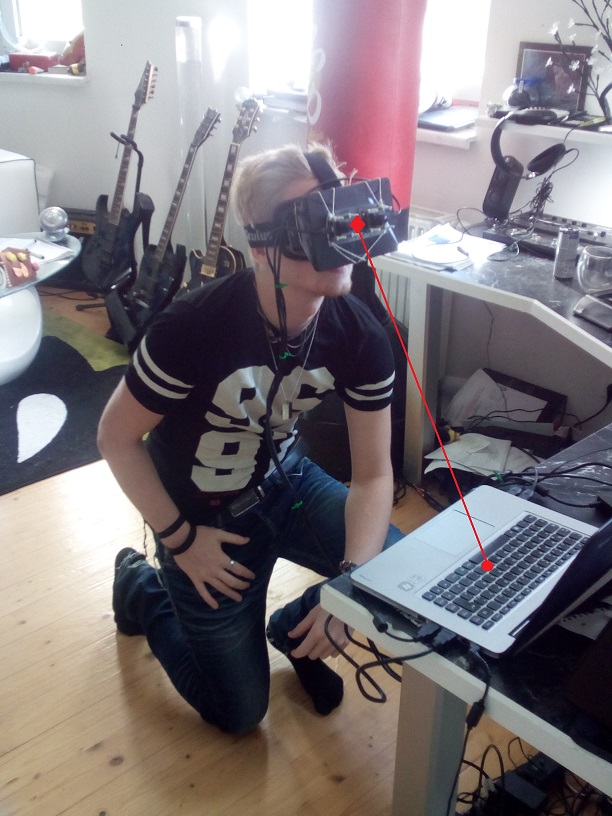}
    \end{subfigure}
    ~ 
    \begin{subfigure}{0.55\textwidth}
        \includegraphics[width=\textwidth]{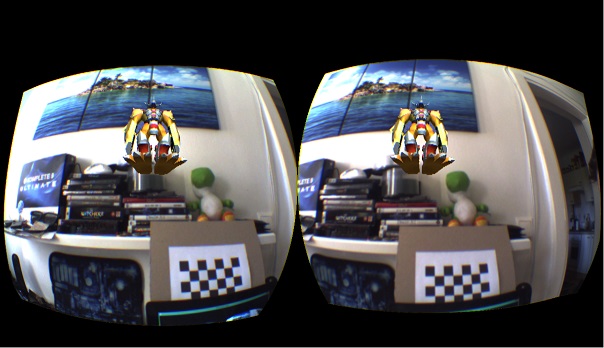}
    \end{subfigure}
    \caption{Results: \textbf{crouching}, near distance. Image property of Markus H\"oll }
\end{figure}
\label{fig:crouch}

 \begin{figure}[H]
    \centering
    \label{fig:far}
    \begin{subfigure}{0.4\textwidth}
        \includegraphics[width=\textwidth]{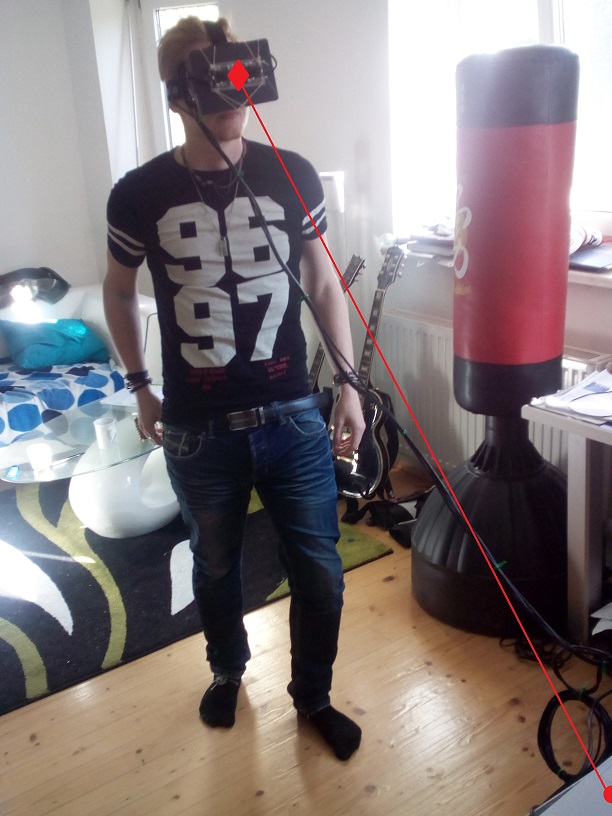}
    \end{subfigure}
    ~ 
    \begin{subfigure}{0.55\textwidth}
        \includegraphics[width=\textwidth]{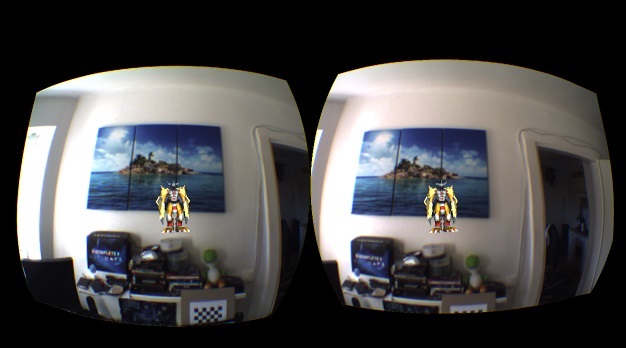}
    \end{subfigure}
    \caption{Results: \textbf{far} distance. Image property of Markus H\"oll }
\end{figure}
\label{fig:far}

 \begin{figure}[H]
    \centering
    \label{fig:side}
    \begin{subfigure}{0.4\textwidth}
        \includegraphics[width=\textwidth]{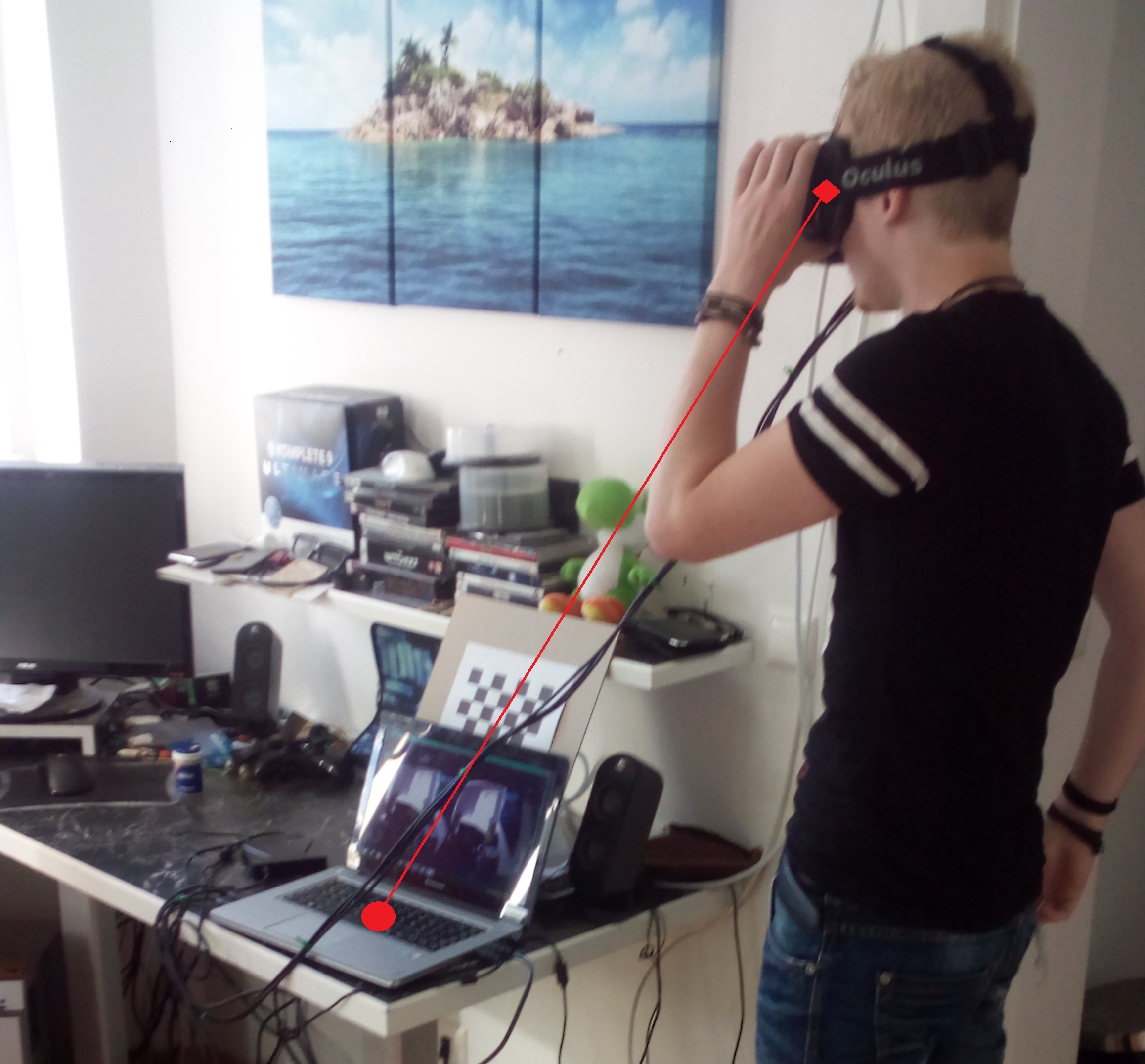}
    \end{subfigure}
    ~ 
    \begin{subfigure}{0.55\textwidth}
        \includegraphics[width=\textwidth]{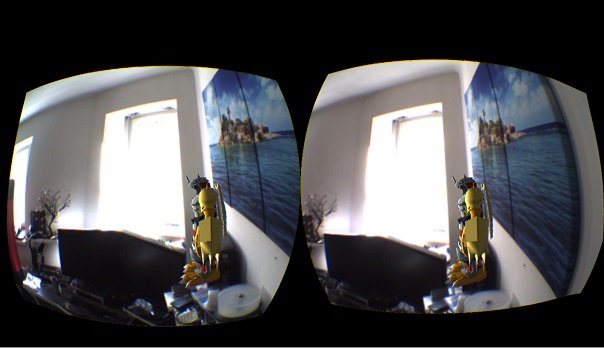}
    \end{subfigure}
    \caption{Results: \textbf{side} view, walking around. Image property of Markus H\"oll }
\end{figure}
\label{fig:side}

\newpage

\section{Future Work}\label{sec:future}

We want to focus now on the interaction with virtual objects. The motivation is  an interaction system in AR which feels more natural like using our hands and interacting with virtual objects the same way or at least similar as we do with real objects. We expect that to increase the immersion of AR significantly. However, we strongly consider a hardware upgrade since the Oculus Rift DK1 is pretty old already and the display's resolution is very low. We will use the HoloLens for this project and see how it feels like if we change the interaction system of it with our own one. It will be very interesting to see how that works out and if we might also use some functionality of the interaction from the HoloLens, like maybe using the ''Gaze'' raycaster for highlighting the virtual objects but further interacting with our hand gestures or totally bypass the whole interaction and only use the new one.

To capture our hand skeleton, we will use the Leap Motion controller which is using structured light for measurement. The Leap Motion has already a VR mount to install it on the front plate of VR HMDs. Since that could be tricky on the HoloLens, a custom mount might be the better choice. What might be more tricky is the communication between the HoloLens and the Leap Motion, since we will need to set up a remote communication between them due to the fact that the USB ports of the HoloLens can not be used for periphery devices appearantly. Another challenge will be to translate the coordinates from the Leap Motion's coordinate system to the AR space of the HoloLens and compute a fine hand gesture detection and collision resolution on virtual objects.

\clearpage
\bibliography{library}
\bibliographystyle{ieeetr}
\appendix

\end{document}